\documentclass[sigconf]{acmart}

\AtBeginDocument{%
  \providecommand\BibTeX{{%
    \normalfont B\kern-0.5em{\scshape i\kern-0.25em b}\kern-0.8em\TeX}}}

\usepackage{subfigure}
\usepackage{color}
\usepackage{bbm}
\usepackage{multirow}
\usepackage{tablefootnote}
\usepackage{balance}

\copyrightyear{2023}
\acmYear{2023}
\setcopyright{acmlicensed}\acmConference[MM '23]{Proceedings of the 31st ACM International Conference on Multimedia}{October 29-November 3, 2023}{Ottawa, ON, Canada}
\acmBooktitle{Proceedings of the 31st ACM International Conference on Multimedia (MM '23), October 29-November 3, 2023, Ottawa, ON, Canada}
\acmPrice{15.00}
\acmDOI{10.1145/3581783.3611721}
\acmISBN{979-8-4007-0108-5/23/10}
\acmSubmissionID{137}

\begin{document}
% \fancyhead{}
\title{Suspected Object Matters: Rethinking Model's Prediction
for One-stage Visual Grounding}

\author{Yang Jiao}
\authornote{Equal contribution.}
\affiliation{\institution{Shanghai Key Lab of Intell. Info. Processing, School of CS, \\Fudan University}
\city{Shanghai}
\country{China}}
\email{yjiao23@m.fudan.edu.cn}
\author{Zequn Jie}
\authornotemark[1]
\affiliation{\institution{Meituan}
\city{Beijing}
\country{China}}
\email{zequn.nus@gmail.com}
\author{Jingjing Chen}
\authornote{Corresponding author.}
\affiliation{\institution{Shanghai Key Lab of Intell. Info. Processing, School of CS, \\Fudan University}
\city{Shanghai}
\country{China}}
\email{chenjingjing@fudan.edu.cn}
\author{Lin Ma}
\authornotemark[2]
\affiliation{\institution{Meituan}
\city{Beijing}
\country{China}}
\email{forest.linma@gmail.com}
\author{Yu-Gang Jiang}
\affiliation{\institution{Shanghai Key Lab of Intell. Info. Processing, School of CS,\\ Fudan University}
\city{Shanghai}
\country{China}}
\email{ygj@fudan.edu.cn}

% \author{G. Tobin}
% \additionalaffiliation{%
% \institution{The Th{\o}rv{\"a}ld Group}
% \streetaddress{1 Th{\o}rv{\"a}ld Circle}
% \city{Hekla}
% \country{Iceland}}
% \affiliation{%
% \institution{Institute for Clarity in Documentation}
% \streetaddress{P.O. Box 1212}
% \state{Ohio}}


% 1. visual cues deficiency --> loss of visual information; 2. expand/enlarge resolution表述不专业，换成multi-resolution feature/information fusion; 3. Knowledge用于指代先验知识，在我们的setting里面不太合适
\begin{abstract}
Recently, one-stage visual grounders attract high attention due to their comparable accuracy but significantly higher efficiency than two-stage grounders. However, inter-object relation modeling has not been well studied for one-stage grounders. Inter-object relationship modeling, though important, is not necessarily performed among all objects, as only part of them are related to the text query and may confuse the model. We call these objects \textit{``suspected objects"}. However, exploring their relationships in the one-stage paradigm is non-trivial because: (1) no object proposals are available as the basis on which to select suspected objects and perform relationship modeling; (2) suspected objects are more confusing than others, as they may share similar semantics, be entangled with certain relationships, etc, and thereby more easily mislead the model's prediction. Toward this end, we propose a Suspected Object Transformation mechanism (SOT), which can be seamlessly integrated into existing CNN and Transformer-based one-stage visual grounders to encourage the target object selection among the suspected ones. Suspected objects are dynamically discovered from a learned activation map adapted to the model’s current discrimination ability during training. Afterward, on top of suspected objects, a Keyword-Aware Discrimination module (KAD) and an Exploration by Random Connection strategy (ERC) are concurrently proposed to help the model rethink its initial prediction. On the one hand, KAD leverages keywords contributing high to suspected object discrimination. On the other hand, ERC allows the model to seek the correct object instead of being trapped in a situation that always exploits the current false prediction. Extensive experiments demonstrate the effectiveness of our proposed method.
\end{abstract}



\maketitle

\section{Introduction}
The Visual Grounding (VG) task~\cite{DBLP:conf/ijcai/YuYXZ0T18,DBLP:conf/iccv/LiuZZW19,DBLP:conf/aaai/LiuWZH20,yang2019fast,yang2020improving,qiao2020referring,qi2020reverie,huang2021look,deng2021transvg,liao2022progressive,sun2021discriminative} aims to detect the specific entity in an image referred by a given referring expression. As such, VG can facilitate users to freely manipulate detection results by the text, which has widespread application prospects in interactive image-editing~\cite{DBLP:journals/vc/PriceB06,DBLP:journals/corr/abs-1904-02225}, cross-modal retrieval~\cite{song2021spatial,DBLP:conf/cvpr/ZhuNCH19}, and so on. The most straightforward approach is a two-stage paradigm~\cite{DBLP:conf/cvpr/HuXRFSD16,DBLP:conf/eccv/YuPYBB16,DBLP:conf/cvpr/00150CSGH19,DBLP:conf/iccv/LiuZZW19,DBLP:conf/iccv/YangLY19,yang2020graph-structured,DBLP:conf/aaai/LiuWZH20}, where several region proposals are first extracted with an off-the-shelf detector and serve as candidates for the subsequent ranking region-expression pairs. However, such a two-stage paradigm is deficient in two aspects: (i) Region proposals generation brings huge computation cost; (ii) Generated proposals only reflect the confidence of the language-agnostic object detector, thus resulting in missing referred objects. To tackle the above problems, the one-stage visual grounding paradigm~\cite{DBLP:journals/corr/abs-1812-03426,yang2019fast,DBLP:conf/cvpr/LiaoLLWCQL20,yang2020improving,huang2021look,deng2021transvg,zhu2022seqtr} has gained great interest in recent years. Motivated by the one-stage detectors~\cite{DBLP:conf/cvpr/RedmonDGF16,DBLP:conf/eccv/LiuAESRFB16,DBLP:conf/iccv/TianSCH19}, one-stage grounding approaches~\cite{DBLP:journals/corr/abs-1812-03426,yang2019fast,zhu2022seqtr} fuses text query embedding with visual features densely at all spatial locations, and directly make the prediction. Such a pipeline gets rid of the heavy region proposal generation in a two-stage pipeline, and thereby avoids valuable regions being filtered~\cite{yang2019fast}.
  
\begin{figure}[t]
    
    \centering
    \includegraphics[width=0.8\linewidth]{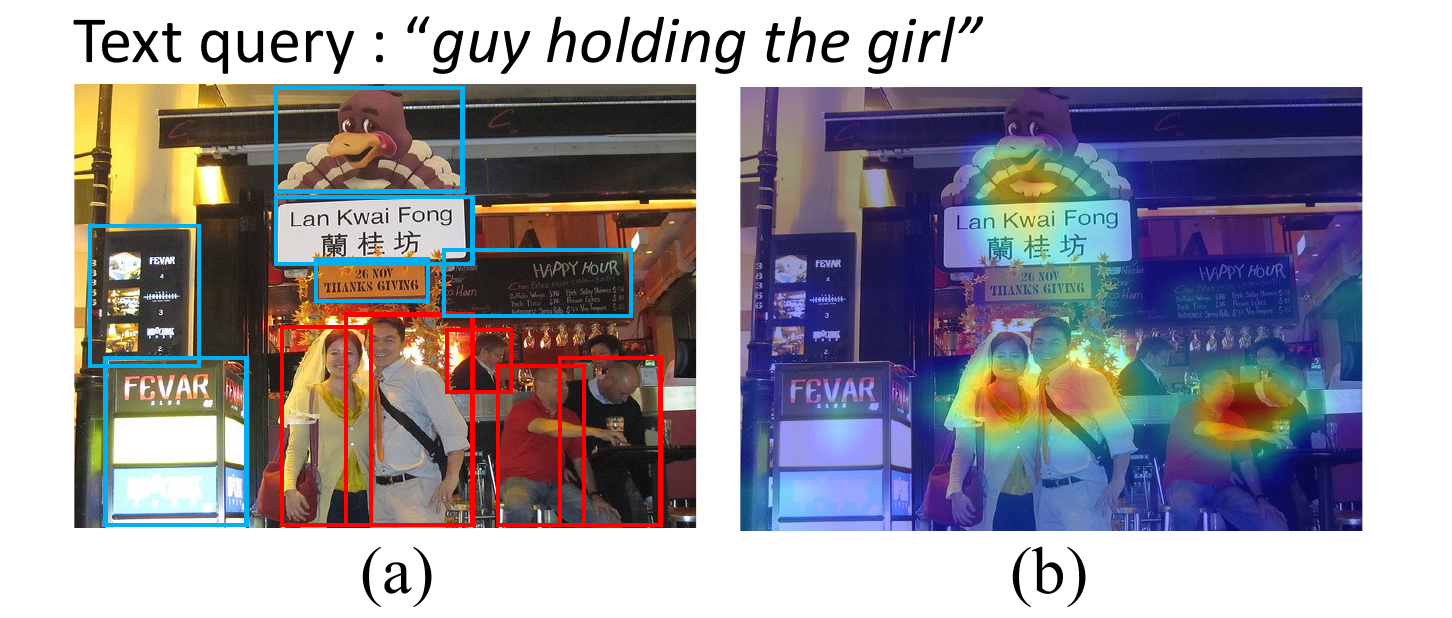}
    \caption{Illustration of (a) text query related objects (inside red boxes) and irrelevant ones (inside blue boxes) according to the given text query; and (b) the activation map generated by a naive one-stage grounder after training five epochs.}
    \label{fig1}
    
\end{figure}
  
\begin{figure}
    \centering
    \includegraphics[width=\linewidth]{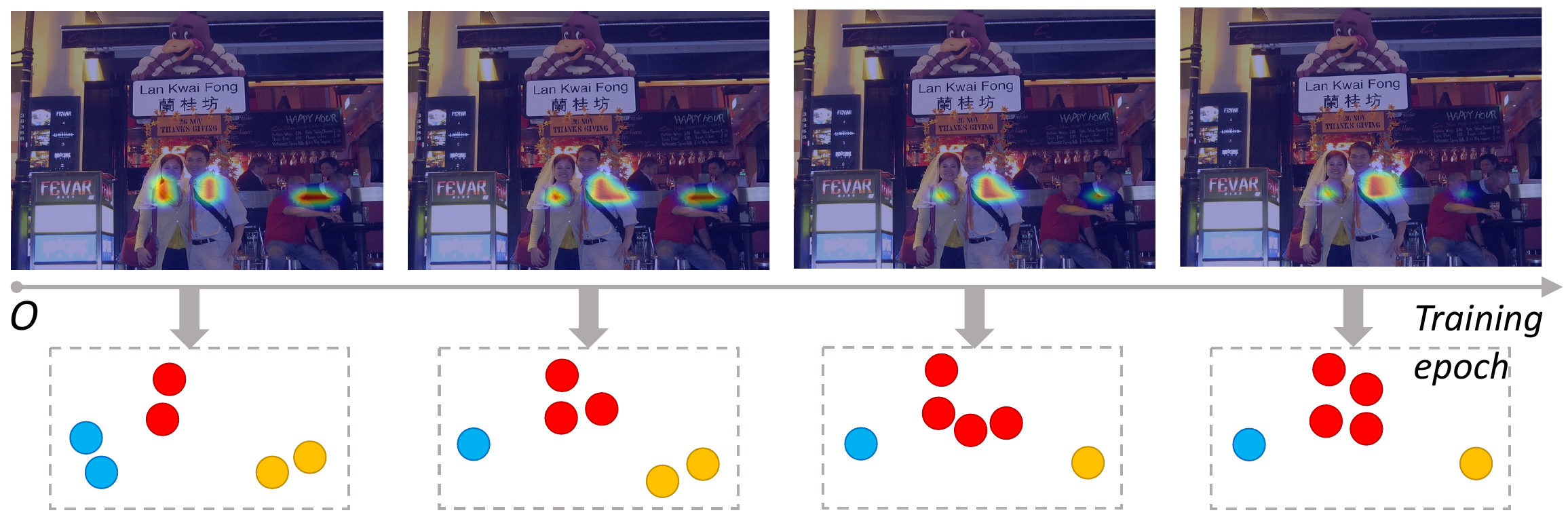}
    \caption{Illustration of the evolution of suspected objects selection during training, where the selected ones gradually converge to the correct object. Please note that the circles with the same color are located in the same neighborhood in the heat map.}
    \label{fig2}
    % \vspace{-14pt}
\end{figure}

Although effective, these one-stage approaches~\cite{DBLP:journals/corr/abs-1812-03426,yang2019fast,huang2021look,deng2021transvg} lack comprehension of relationships between objects, especially those hard to distinguish, which is of vital importance to ground the correct object. For example, as shown in Fig.\ref{fig1}(a), investigating relationships among objects inside all labeled bounding boxes is beneficial for building accurate cross-modal correspondence, however, not all objects are worth being equally treated, as only the objects (\textit{i.e.}, men and a woman) inside red bounding boxes are related to the text query \textit{``guy holding the girl"}. We call these objects as \textbf{\textit{``suspected objects"}} in this paper. Compared with other objects, relationships between these suspected objects should be further investigated to identify more distinctive clues for the correct grounding. 

Several two-stage methods~\cite{DBLP:conf/cvpr/00150CSGH19,DBLP:conf/iccv/YangLY19,DBLP:conf/aaai/LiuWZH20,yang2020graph-structured,chen2021ref} have studied the inter-object relationship modeling problem. They share the idea of constructing a graph with object proposals denoted as nodes and their relations as edges to benefit learning the correspondence between objects and the referring sentence. Among them, LGRAN~\cite{DBLP:conf/cvpr/00150CSGH19} considers intra-class and inter-class relationships for edge representations. SGMN~\cite{yang2020graph-structured} makes the fine-grained classification on each edge according to relative distance, relative angle, and Intersection-over-Union of the connected object proposals. Ref-NMS~\cite{chen2021ref} improves the quality of graph nodes by measuring the relevance between each region candidate and the referring sentence when performing the NMS. Though effective enough, such graphs are designed inherently for two-stage methods, requiring object proposals to be extracted as nodes at the first stage. Since the key improvement of the one-stage paradigm lies in getting rid of the problematic region proposal generation process, the graph design in two-stage methods cannot be directly transferred to the one-stage framework~\cite{DBLP:journals/corr/abs-1812-03426,yang2019fast,huang2021look,deng2021transvg}. The most straightforward way to realize inter-object communication in the one-stage grounding paradigm is to enlarge the receptive field via conventional non-local~\cite{DBLP:conf/cvpr/0004GGH18} operation, deformable convolution~\cite{DBLP:conf/iccv/DaiQXLZHW17}, etc.
Nevertheless, such methods conduct exhaustive message passing for every grid, which is redundant and may even introduce noise from cluttered background and objects irrelevant to the text query, thus hurting the model's performance as illustrated in Table.\ref{abl:graph_type}. In this paper, aiming at concise and effective inter-object relationships modeling under the umbrella of one-stage visual grounding, we propose a novel Suspected Object Transformation mechanism (SOT), which dynamically discovers suspected objects relevant to the text query and gradually disambiguates the referred one from them.

% The very first core problem is how to obtain the suspected objects relevant to the text query as graph nodes without the aid of object proposals.
The primary problem is how to discover suspected objects while preserving the complete scene for further exploration.
Fortunately, we observe that even a naive one-stage grounder can pay high attention to only a limited number of objects after training a few epochs as shown in Fig.\ref{fig1}(b). Meanwhile, these objects are highly consistent with the suspected objects labeled by red bounding boxes in Fig.\ref{fig1}(a)\footnote{More relevant visualization results are included in the supplementary materials.}.
Hence, suspected objects can be selected according to the one-stage grounder's initial confidence. It is worth mentioning that unlike the fixed region proposals in the two-stage methods~\cite{DBLP:conf/cvpr/00150CSGH19,DBLP:conf/iccv/YangLY19,DBLP:conf/aaai/LiuWZH20}, the selected suspected objects can dynamically evolve and finally converge to the correct object during training as illustrated in Fig.\ref{fig2}, demonstrating the progressively enhanced suspected object selection ability of the proposed SOT.
However, the model's initial confidence can be unreliable, as these suspected objects are inherently more confusing than others. As shown in Fig.\ref{fig1}, they may share similar semantics (several men can correspond to the \textit{``guy"} referred in the expression), entangled within a relationship (the relation \textit{``holding"} binds the guy and girl), etc. Therefore, aiming to help the model rethink its initial selection, we propose a Keyword-Aware Discrimination module (KAD) and an Exploration by Random Connection strategy (ERC) within the SOT. On the one hand, KAD leverages linguistic keywords contributing high to suspected objects discrimination. On the other hand, ERC allows the model to seek the correct object instead of being trapped in a situation that always exploits the current false prediction.

In summary, our contributions lie in the following threefold. (1) We explore the inter-object relationships under the umbrella of the one-stage visual grounding framework and propose a Suspected Object Transformation mechanism (SOT) to disambiguate the referred object from multiple suspected objects. (2) A novel Keyword-Aware Discrimination (KAD) module, as well as an Exploration by Random Connection strategy (ERC), are proposed within SOT to facilitate language-guided suspected object discrimination and correct object exploration, respectively. (3) Extensive experimental results on prevalent visual grounding benchmarks demonstrate the effectiveness of our method.

\begin{figure*}[!t]
  \centering
  \includegraphics[width=\linewidth]{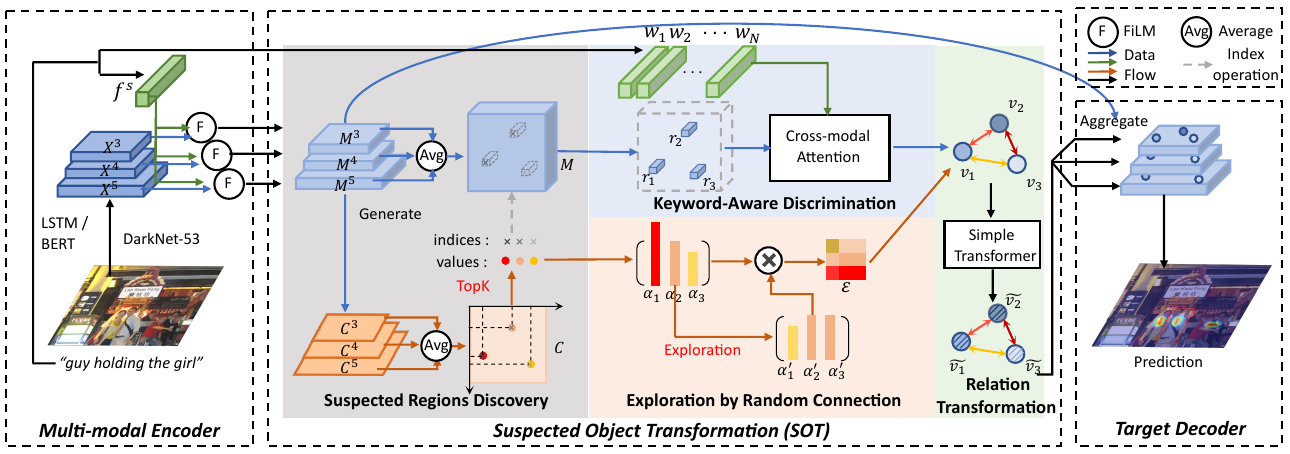}
   \caption{The overall framework of our proposed Suspected Object Transformation (SOT) module with a CNN-based one-stage grounder. The visual and textual features are first extracted and interacted in the multi-modal encoder to generate multi-modal features, which are then fed into SOT to select multiple suspected objects. These suspected objects features are updated by re-evaluating their relations to keywords and exploring their inter-relations to identify more distinctive clues for distinguishing the referred one from them. Finally, the updated suspected object features are aggregated with the multi-modal features and make the final prediction. To keep the picture concise, we only illustrate the situation when $\textit{K}$ equals 3. ``$\otimes$" represents the outer product.}
   \label{framework}
%   \vspace{-16pt}
\end{figure*}

%------------------------------------------------------------------------
% \vspace{-4pt}
\section{Related Work}

%-------------------------------------------------------------------------
%\subsection{Referring Image Segmentation}
\label{sec:related_work}
% \vspace{-4pt}
\subsection{Two-stage Visual Grounding.} Enjoying the benefits of several off-the-shelf two-stage object detectors~\cite{ren2015faster,DBLP:conf/iccv/HeGDG17}, candidate instances in the image are well-extracted, and therefore existing two-stage visual grounding approaches~\cite{DBLP:conf/cvpr/HuXRFSD16,DBLP:conf/eccv/YuPYBB16,DBLP:conf/cvpr/00150CSGH19,DBLP:conf/iccv/LiuZZW19,DBLP:conf/iccv/YangLY19,yang2020graph-structured,DBLP:conf/aaai/LiuWZH20} mainly make efforts on devising effective matching mechanism between each candidate and referring expression. In the early years, Hu \textit{et al.}~\cite{DBLP:conf/cvpr/HuXRFSD16} synthesized global context, local descriptor as well as spatial configuration for candidates when scoring with text queries, while Yu \textit{et al.}~\cite{DBLP:conf/eccv/YuPYBB16} measures visual differences between multiple candidates. Furthermore, inspired by the success of graph neural networks~\cite{DBLP:conf/iclr/KipfW17,DBLP:journals/corr/abs-1710-10903}, various works exploit attributes and relationships of objects~\cite{DBLP:conf/cvpr/00150CSGH19}, cross-modal interactions~\cite{DBLP:conf/aaai/LiuWZH20}, reasoning techniques~\cite{DBLP:conf/iccv/YangLY19,jiao2022more,li-etal-2023-multi-modal,li-etal-2023-neural}, etc. based on a graph-structured framework. However, the above modeling over-simplifies the composite nature of language into a holistic sentence embedding or a rough subject-predicate object triplet. To this end, Liu~\cite{DBLP:conf/iccv/LiuZZW19} \textit{et al.} designs a neural modular tree network to ground each language composite into an image along the dependency parser tree of the sentence. Similarly, Yang \textit{et al.}~\cite{yang2020graph-structured} parse the sentence into a language scene graph and conduct structured reasoning. Recently, reasoning through interaction with large language models~\cite{li2023lmeye} has become a trend in this field.

% \vspace{-8pt}
\subsection{One-stage Visual Grounding.}  Recently, the one-stage visual grounding has gained great interest, as it solves two limitations of the two-stage paradigm, \textit{i.e.} the performance cap caused by inaccurate object proposals results, and the long inference time caused by heavy computation in generating object proposals. Instead of extracting all candidate objects in the image, one-stage grounding approaches~\cite{DBLP:journals/corr/abs-1812-03426,yang2019fast,DBLP:conf/cvpr/LiaoLLWCQL20,yang2020improving,huang2021look,deng2021transvg} formulate the whole grounding process in an end-to-end manner. As the pioneer works, SSG~\cite{DBLP:journals/corr/abs-1812-03426} and FAQA~\cite{yang2019fast} densely fuse the sentence feature with image features at every locations, and directly make the final prediction. Later, RCCF~\cite{DBLP:conf/cvpr/LiaoLLWCQL20} proposes a cross-modal correlation filter mechanism to select the peak response area as prediction results. Afterward, researchers further make efforts in modeling visual inter-object relations and linguistic contexts. To promote region interactions, the LBYLNet~\cite{huang2021look} investigates an efficient landmark convolution. Moreover, considering the composite nature of language, ReSC~\cite{yang2020improving} recursively generates sub-queries from sentences to interact with visual features for multiple rounds. Recently, Transformer-based grounders~\cite{deng2021transvg,li2021referring} have proposed for fine-grained vision-language communication. Among them, Seq-TR~\cite{zhu2022seqtr} re-formulates the prediction as a points sequence regression process, which effectively addresses a wide range of tasks, including visual grounding and referring image segmentation~\cite{jiao2021two}. Although effective, these methods treat each patch equally without explicitly excavating object-level relationships.

% %------------------------------------------------------------------------
\section{Method}
\label{sec:methods}
% \vspace{-8pt}
Generally, a one-stage visual grounder consists of a multi-modal encoder and a target decoder, where the former extract uni-modal features and interact them to generate multi-modal ones, and the latter decodes the multi-modal features to generate predictions. To explicitly present the complete workflow of a one-stage grounder with our Suspected Object Transformation (SOT) mechanism, we take the CNN-based grounder as an example in the following. And combining our SOT with Transformer-based grounders will be briefly introduced in Sec.\ref{SOT_with_Transformer}.  

As shown in Fig.\ref{framework}, we insert the SOT between the multi-modal encoder and the target decoder. With the features obtained from the multi-modal encoder, SOT can dynamically identify the suspected objects and explore their relationships. Afterward, the resulting suspected object features together with the multimodal features are consumed by the target decoder for the final prediction.

% The overall framework of our proposed method is shown in Fig.\ref{framework}, where the three components, namely multi-modal encoder, suspected object transformation block, and prediction are coupled together and can be trained in an end-to-end fashion. Given a paired image and referring expression as input, the multi-modal encoder extracts the visual and textual features and performs cross-modal interactions to generate multi-modal features, which are then fed into the Suspected Object Transformation (SOT) block to dynamically identify the suspected objects and exploit their relationships. Afterward, the updated node features of SOT, mainly characterizing the suspected objects, are jointly considered  with the multi-modal features to make the final grounding prediction.

\subsection{Multi-modal Encoder}
\label{subsec:multi-modal_encoder}
As shown in Fig.\ref{framework}, the multi-modal takes an image and a text query as inputs.
The sentence feature $f^{s}$ and word features $\{w_{n}\}_{n=1}^{N}$ ($N$ is the sentence length) are extracted with an LSTM~\cite{hochreiter1997long} or a BERT~\cite{DBLP:conf/naacl/DevlinCLT19} encoder. Multi-level visual features of the image are extracted with a DarkNet-53~\cite{DBLP:conf/cvpr/LinDGHHB17} backbone, and then mapped to the same channel dimension using a 1$\times$1 convolution. And each level of visual features are concatenated with an 8-D coordinate map to generate position-aware visual features $X^{l} \in \mathbb{R}^{h_{l}\times w_{l}\times d_{m}}$, where $h_{l}$ and $w_{l}$ are the height and width dimension of $l$-th level of visual feature ($l=\{3,4,5\}$ in our implementation). Afterward,  FiLM~\cite{DBLP:conf/aaai/PerezSVDC18} is used to  modulate $X^{l}$ with the sentence feature $f^{s}$ as:
\begin{equation}
    \begin{split}
        % \gamma^{l} & = \mathrm{FC}^{l}_{\gamma}(f^{s}),\quad \beta^{l} = \mathrm{FC}^{l}_{\beta}(f^{s}), \\
        % M^{l} &  = \mathrm{FiLM}(X^{l}|\gamma^{l},\beta^{l}) = \gamma^{l} \odot X^{l} \oplus \beta^{l},
        M^{l} & = \mathrm{FiLM}(X^{l},f^{s}) \\
              & = \mathrm{FC}^{l}_{\gamma}(f^{s}) \odot X^{l} \oplus \mathrm{FC}^{l}_{\beta}(f^{s})
    \end{split}
\label{eq:1}
\end{equation}
where $\mathrm{FC}^{l}_{\gamma}(\cdot)$ and $\mathrm{FC}^{l}_{\beta}(\cdot)$ are two fully-connected layers. ``$\odot$'' and ``$\oplus$'' represent element-wise multiplication and addition operations with broadcasting, respectively. Through the FiLM operation, grid features at different locations in multi-modal feature $M^{l}$ are adaptively activated as the response to the text query. Different levels of multi-modal features $M^{3},M^{4},M^{5}$ are then served as inputs for SOT.
% visual feature's activations are modulated by coefficient vectors $\gamma^{l}$ and $\beta^{l}$ generated from the textual feature, resulting the multi-modal feature $M^{l}\in \mathbb{R}^{h_{l}\times w_{l}\times d_{m}}$, which will serve as input for our proposed SOG.   
% LSTM/BERT~\cite{DBLP:conf/naacl/DevlinCLT19} 
% \vspace{-12pt}
\subsection{Suspected Object Transformation}
\label{subsec:suspected_object_graph}
% \vspace{-8pt}
As the core component of our method, SOT is responsible for selecting the suspected objects and updating their feature representations, so as to facilitate the model to rethink and gradually correct its selection. %We denote SOG as $\mathcal{G}=(\mathcal{V,E})$, with $\mathcal{V}$ and $\mathcal{E}$ being nodes and edges, respectively.
As illustrated in Fig.\ref{framework}, the whole SOT consists of four stages. In the first stage, we select several grids with high activation scores based on multi-modal features $M^{3},M^{4},M^{5}$, and regard them as the suspected regions. In the second stage, these suspected regions are discriminated by re-evaluating their relations to keywords. Subsequently, in the third stage, the activation scores corresponding to these suspected regions are adjusted by exploring new connection strengthens to learn more comprehensive intra-object relations. Finally, the updated suspected objects and their relations are communicated with a simple transformer block.
% the corresponding design motivation as well as implementation details of which are introduced in the following subsections.

% \vspace{-16pt}
\subsubsection{Suspected Regions Discovery.}
\label{subsubsec:suspected_regions_discovery}
\quad Considering that no off-the-shelf object candidates are available in one-stage VG, we first identify the regions with suspected objects on which inter-object relationships are expected to be built. Multi-modal feature $M^{l}$ shows the relevance to the text query in each visual grid. Hence, to facilitate the subsequent efficient computation, a low dimensional textual activation map $C^{l}\in \mathbb{R}^{h_{l}\times w_{l}\times 1}$ is produced by dimension reduction with a simple convolution layer:
%applied to $M^{l}$.  
% \blue{As grids in multi-modal feature $M^{l}$ are highlighted or suppressed according to their relevance to the textual cues. 
% we can directly measure the activation degree of each grid by reducing the channel dimension of $M^{l}$ to generate an activation map $C^{l}\in \mathbb{R}^{h_{l}\times w_{l}\times 1}$}
% we reduce the channel dimension of $M^{l}$ and obtain an attention map $C^{l}\in \mathbb{R}^{h_{l}\times w_{l}\times 1}$, which denotes
% % apply simple dimension reduction transformations on $M^{l}$ and generate a 1-channel map $C^{l}\in \mathbb{R}^{h_{l}\times w_{l}\times 1}$, denoting
% the confidence that referred object falls on each grid:}
% \textcolor{blue}{model's interest} to every grids:
\begin{equation}
    C^{l} = \mathrm{ReLU}(\mathrm{Conv}(M^{l})).
\end{equation}
To discover suspected objects with different sizes, we integrate $C^{l}$ from different scales (\textit{i.e.}, $C_{3},C_{4},C_{5}$) to obtain the multi-scale activation map $\overline{C}$ with average operation. The assembled multi-modal feature $\overline{M}$ is also generated in a similar way:

% Separately selecting the most intensely activated grids inside each level of $C^{l}$ lacks cross-level comparison and may hurt the accuracy of locating suspected objects, as different scales of objects are generally activated in different levels of confidence map $C^{l}$.
% % , and such selection behavior \textcolor{blue}{lacks global comparison}. 
% Therefore, we assemble $C_{3},C_{4},C_{5}$ to obtain a global activation map $\overline{C}$ with average operation, and the assembled multi-modal feature $\overline{M}$ is also generated the same way:
\begin{equation}
    \overline{M} = \frac{1}{3} \sum_{l=3}^{5}M^{l}, \quad \overline{C} = \frac{1}{3} \sum_{l=3}^{5}C^{l},
\end{equation}
where we omit the upsample and downsample operations on $M^{3}$ and $M^{5}$. 

Based on the obtained multi-modal feature $\overline{M}$ and activation map $\overline{C}$, we choose $\textit{K}$ ($\textit{K}$=$6$ in our implementation) grid features from $\overline{M}$ with the largest intensities in $\overline{C}$ as the suspected regions. We denote the $k$th largest activation score as $\alpha_{k}$, and the corresponding grid feature (\textit{i.e.,} suspected region) as $r_{k}$. By dynamically selecting $\textit{K}$ most intensively activated grid features $\{r_{k}\}_{k=1}^{K}$ at each training epoch, the selected grids of SOT accordingly evolve along with the training proceeds, such that the model can gradually focus on the most confusing suspected objects to facilitate the correct prediction.

\begin{figure}[!t]
  \centering
  \includegraphics[width=0.8\linewidth]{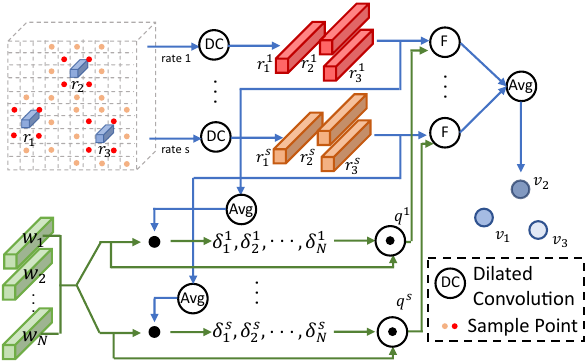}
  \caption{The detailed design of the cross-modal attention module in keyword-aware node representation, where “$\cdot$" represents the dot product operation, and “$\odot$" stands for the element-wise multiplication with broadcasting.}
  \label{figure_KC}
  % \vspace{-8pt}
\end{figure}

\subsubsection{Keyword-aware Discrimination.}
\label{subsubsec:KAD}
\quad As illustrated in Fig.\ref{fig1}, the suspected objects (\textit{e.g.,} the guy, girl, and sitting man) are usually visually similar and require informative keywords (\textit{e.g.,} holding) in the text query to be further distinguished. Inspired by this, we propose a cross-modal attention module to learn the importance of each word in the referred object discrimination and assemble word representations weighted by the importance scores to modulate the suspected regions $r_{k}$ as shown in Fig.\ref{figure_KC}.
% \textcolor{blue}{construct the keyword query} from \textcolor{blue}{word features $w_{n}$ of the referring expression} and modulate suspected regions $r_{k}$ obtained from the last step as shown in Fig.~\ref{figure_KC}. 

Instead of directly utilizing grid features to interact with word features, we first incorporate context information from the neighboring regions for every suspected region $r_{k}$ to effectively learn the representation for objects with different sizes.  We apply 3$\times$3 dilated convolution~\cite{DBLP:journals/corr/YuK15} with dilation rate $s$ to aggregate richer object information for every $r_{k}$, and obtain corresponding context-aware suspected region features $\{r_{k}^{s}\}_{k=1}^{K}$ accordingly:
% \vspace{-6pt}
\begin{equation}
    r_{k}^{s} = \sum_{u}(W_{u} g_{u}), \quad \forall u \in \mathcal{N}^{s}(r_{k}),
    % \vspace{-8pt}
\end{equation}
% \vspace{-4pt}
where $\mathcal{N}^{s}(r_{k})$ is the neighborhood of $r_{k}$ (including the $r_{k}$) when dilation rate is $s$, and $g_{u}$ is a grid feature in this neighborhood. Afterward, $\{r_{k}^{s}\}_{k=1}^{K}$ are averaged and matched with every word feature $w_{n}$ to generate the importance score $\delta_{n}^{s}$:
% \vspace{-6pt}
\begin{equation}
    \begin{split}
        & \overline{r}^{s} = \frac{1}{K}\sum_{k=1}^{K}r_{k}^{s}, \\
        & \delta_{n}^{s} = \frac{\mathrm{exp}(\overline{r}^{s} \cdot w_{n})}{\sum_{m=1}^{N}\mathrm{exp}(\overline{r}^{s} \cdot w_{m})},
    \end{split}
\end{equation}
% \vspace{-2pt}
where ``$\cdot$'' denotes the dot product. Then we calculate a keyword-aware textual representation $q^{s}$ with a weighted combination of word features, where the word with higher importance score $\delta_{n}^{s}$ contributes more in constructing $q^{s}$. 
Next, the region feature $r_{k}^{s}$ is first modulated by the keyword-aware textual representation $q^{s}$ with the FiLM module, and then aggregated to generate the corresponding suspected object feature $v_{k}$. The above operations can be formulated as:
\begin{equation}
    \begin{split}
        & q^{s} = \sum_{n=1}^{N}\delta_{n}^{s} \odot w_{n}, \\
        & v_{k} = \frac{1}{|\mathcal{S}|}\sum_{s}\mathrm{FiLM}(r_{k}^{s},q^{s}), \quad \forall s \in \mathcal{S},
    \end{split}
\end{equation}
where $\mathcal{S}$ is the collection of values for the dilation rate ($\mathcal{S}$=$\{1,6,12\}$ in our implementation). We denote the set of suspected objects feature representations as $\mathcal{V}=\{v_{k}\}_{k=1}^{K}$.

% \vspace{-12pt}
\subsubsection{Exploration by Random Connection.}
\label{subsubsec:ERC}
\begin{figure}[!t]
  \centering
 \includegraphics[width=\linewidth]{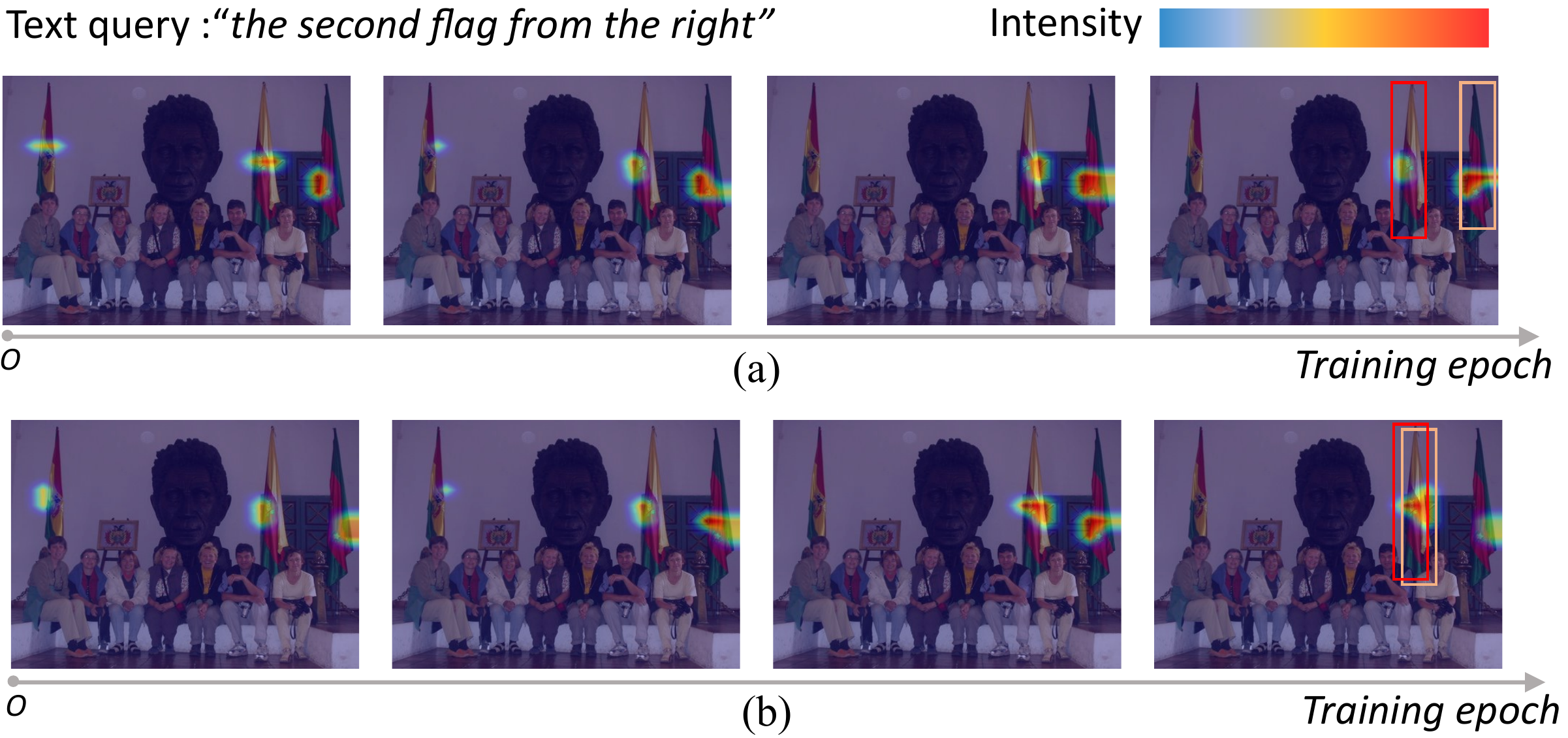}
\caption{The influence of our proposed exploration by random connection strategy (ERC) to suspected object selection evolution process. The red and orange bounding boxs represent the ground-truth and model prediction results, respectively. (a) Without the ERC, the previous false judgment is passed on and accumulated with training proceeds, demonstrating the error remaining problem. (b) With the ERC, inter-object relation strengths are decoupled from model judgment confidence, and therefore prevent the wrong judgment from affecting the later decision, which finally yields the correct grounding results.}
\label{Exploration}
% \vspace{-12pt}
\end{figure}

\quad With $\textit{K}$ activation scores $\{\alpha_{k}\}_{k=1}^{K}$ as inputs, a straightforward way to represent the relation $e_{ij}$ between suspected objects $v_{i}$ and $v_{j}$ can be defined following the query and key calculation in the Transformer~\cite{vaswani2017attention}:
\begin{equation}
    e_{ij} = \alpha_{i} \odot \alpha_{j}.
\label{edge}
\end{equation}
However, the above relation calculation strategy imposes high connection strength between the suspected objects with high activation scores. In this way, an incorrect object with high activation would have a higher impact after inter-object message passing. Hence, if the initial incorrect objects obtain high activation scores, the model is easily trapped in a situation that always exploits the current false object selection as shown in Fig.\ref{Exploration} (a), deviating from the goal of the object selection rethinking and correction. 

To balance between trusting the model's initial judgment and preventing the aforementioned error remaining problem, we propose an \textit{exploration by random connection} strategy to moderately decouple the relation strength from suspected object activations. Specifically, motivated by the data augmentation technique in BERT~\cite{DBLP:conf/naacl/DevlinCLT19}  that replaces a certain token with another random token 10\% of the time, we remain $\alpha_{k}$ unchanged with the probability $p$ ($p=0.5$ in our implementation), and uniformly sample a substitution from other $\alpha$ values. We denote $\alpha'_{k}$ as the result of $\alpha_{k}$ processed with exploration by random connection, and $e_{ij}$ can be reformulated by replacing $\alpha_{j}$ with $\alpha'_{j}$ in Eq.(\ref{edge}) accordingly:
\begin{equation}
    e_{ij} = \alpha_{i} \odot \alpha'_{j}.
\label{edge_exploration}
\end{equation}
As such, the set of explored inter-object relations are obtained, which is formulated as $\mathcal{E} = \{e_{ij}\}_{i,j=1}^{K}$. Then, we further modulate each suspected object representation by aggregating its neighboring ones with the explored relationships with a simple transformer block as:
\begin{equation}
    \widetilde{v_{k}} = \sum_{j \in \{1,2,\dots,K\}} e_{jk} \odot v_{j},
\end{equation}
% where $\widetilde{v_{k}}$ is the final updated  representation.

\subsection{Prediction}
\label{subsec:prediction_head}
% \vspace{-4pt}
Carrying more distinctive clues of the referred object, the updated suspected object features $\{\widetilde{v_{k}}\}_{k=1}^{K}$ are scattered to the corresponding positions of multi-modal features $M^{3},M^{4}$ and $M^{5}$, and then fed into a prediction head for grounding the referred object.

\subsection{SOT with Transformer}
\label{SOT_with_Transformer}
The main differences between Transformer-based and CNN-based one-stage grounders are the architecture of the multi-modal encoder and the prediction target in the target decoder. First, since the transformer block inherently enables the multi-modal interaction, the multi-modal encoder in transformer-based grounders consists of stacked transformer blocks, while CNN-based grounders require a deliberately designed operator to integrate multi-modal information, such as the FiLM module described in Eq(\ref{eq:1}). Second, transformer-based grounders adopt transformer decoder blocks to directly regress box coordinates in an auto-regressive manner, while CNN-based grounders predict the offsets between the anchor box and ground truth following the YOLO detector. Similar to the pipeline introduced in CNN-based grounder, we select SeqTR~\cite{zhu2022seqtr} as our transformer-based baseline, and insert the proposed SOT between its transformer encoder and target decoder. In the suspected regions discovery part, we only use the output of the last transformer encoder block. And the KAD and ERC are the same as introduced in Sec.\ref{subsubsec:KAD} and Sec.\ref{subsubsec:ERC}.

\begin{table*}[!t]
\centering
\caption{Comparison with state-of-the-art methods on RefCOCO, RefCOCO+, RefCOCOg. ``R-101", ``R-50" and ``D-53" represent ``ResNet-101", ``ResNet-50" and "DarkNet-53", respectively. In two-stage methods, we use * to indicate that this model uses ground-truth bounding boxes as input. In transformer-based methods, we use $^\dagger$ to indicate that this model is trained using additional mask annotation. When comparing inference latency, we use $^\star$ to indicate that this latency is evaluated with better hardware, for example, GTX 2080 Ti (RefTrans) vs GTX-1080Ti (ours). The best performances in two-stage methods (except the DGA* for fair comparison) are highlighted with \underline{underline}, and the best performances in transformer-based and CNN-based one-stage methods are highlighted with \textbf{\textit{bold italics}} and \textbf{bold}, respectively.}
\label{SOTA:COCO_series}
\scalebox{1}
{\begin{tabular}{ccccccccccc}
\toprule
\multicolumn{1}{c|}{\multirow{2}{*}{Methods}} & \multicolumn{1}{c|}{\multirow{2}{*}{\begin{tabular}[c]{@{}c@{}}Visual \& Text \\ Encoder\end{tabular}}} & \multicolumn{1}{c|}{\multirow{2}{*}{\begin{tabular}[c]{@{}c@{}}Image\\ size\end{tabular}}} & \multicolumn{3}{c|}{RefCOCO}                                                                                       & \multicolumn{3}{c|}{RefCOCO+}                                                                                      & \multicolumn{1}{c|}{RefCOCOg}                      & \multirow{2}{*}{\begin{tabular}[c]{@{}c@{}}Time\\ (ms)\end{tabular}} \\
\multicolumn{1}{c|}{}                         & \multicolumn{1}{c|}{}                                                                                   & \multicolumn{1}{c|}{}                                                                      & val                           & testA                         & \multicolumn{1}{c|}{testB}                         & val-g                           & testA                         & \multicolumn{1}{c|}{testB}                         & \multicolumn{1}{c|}{val}                           &                                                                      \\ \hline
\multicolumn{11}{c}{\textit{Two-stage methods}}                                                                                                                                                                                                                                                                                                                                                                                                                                                                                                                                                                                     \\ \hline
\multicolumn{1}{c|}{CMN~\cite{DBLP:conf/cvpr/HuRADS17}}                      & \multicolumn{1}{c|}{VGG16 \& LSTM}                                                                      & \multicolumn{1}{c|}{-}                                                                     & -                             & 71.03                         & \multicolumn{1}{c|}{65.77}                         & -                             & 54.32                         & \multicolumn{1}{c|}{47.76}                         & \multicolumn{1}{c|}{-}                             & -                                                                    \\
\multicolumn{1}{c|}{VC~\cite{DBLP:conf/cvpr/ZhangNC18}}                       & \multicolumn{1}{c|}{VGG16 \& LSTM}                                                                      & \multicolumn{1}{c|}{-}                                                                     & -                             & 73.33                         & \multicolumn{1}{c|}{67.44}                         & -                             & 50.86                         & \multicolumn{1}{c|}{58.03}                         & \multicolumn{1}{c|}{-}                             & -                                                                    \\
\multicolumn{1}{c|}{ParallelAttn~\cite{DBLP:conf/cvpr/ZhuangWS0H18}}             & \multicolumn{1}{c|}{VGG16 \& LSTM}                                                                      & \multicolumn{1}{c|}{-}                                                                     & -                             & 75.31                         & \multicolumn{1}{c|}{65.52}                         & -                             & 61.34                         & \multicolumn{1}{c|}{50.86}                         & \multicolumn{1}{c|}{-}                             & -                                                                    \\
\multicolumn{1}{c|}{LGRAN~\cite{DBLP:conf/cvpr/00150CSGH19}}                    & \multicolumn{1}{c|}{VGG16 \& LSTM}                                                                      & \multicolumn{1}{c|}{-}                                                                     & -                             & 76.6                          & \multicolumn{1}{c|}{66.4}                          & -                             & 64.00                         & \multicolumn{1}{c|}{53.40}                         & \multicolumn{1}{c|}{61.78}                         & -                                                                    \\
\multicolumn{1}{c|}{SLR~\cite{DBLP:conf/cvpr/YuTBB17}}                      & \multicolumn{1}{c|}{R-101 \& LSTM}                                                                 & \multicolumn{1}{c|}{-}                                                                     & 69.48                         & 73.71                         & \multicolumn{1}{c|}{64.96}                         & 55.71                         & 60.74                         & \multicolumn{1}{c|}{48.80}                         & \multicolumn{1}{c|}{-}                             & -                                                                    \\
\multicolumn{1}{c|}{MAttNet~\cite{DBLP:conf/cvpr/Yu0SYLBB18}}                  & \multicolumn{1}{c|}{R-101 \& LSTM}                                                                 & \multicolumn{1}{c|}{-}                                                                     & 76.40                         & 80.43                         & \multicolumn{1}{c|}{69.28}                         & 64.93                         & 70.26                         & \multicolumn{1}{c|}{56.00}                         & \multicolumn{1}{c|}{-}                             & 320                                                                  \\
\multicolumn{1}{c|}{DGA~\cite{DBLP:conf/iccv/YangLY19}}                      & \multicolumn{1}{c|}{R-101 \& LSTM}                                                                 & \multicolumn{1}{c|}{-}                                                                     & -                             & 78.42                         & \multicolumn{1}{c|}{65.53}                         & -                             & 69.07                         & \multicolumn{1}{c|}{51.99}                         & \multicolumn{1}{c|}{-}                             & 341                                                                  \\
\multicolumn{1}{c|}{DGA*~\cite{DBLP:conf/iccv/YangLY19}}                      & \multicolumn{1}{c|}{R-101 \& LSTM}                                                                 & \multicolumn{1}{c|}{-}                                                                     & 86.34                             & 86.64                         & \multicolumn{1}{c|}{84.79}                         & 73.56                             & 78.31                         & \multicolumn{1}{c|}{68.15}                         & \multicolumn{1}{c|}{80.21}                             & 341                                                                  \\
\multicolumn{1}{c|}{CM-A-E~\cite{DBLP:conf/cvpr/LiuWSWL19}}             & \multicolumn{1}{c|}{R-101 \& LSTM}                                                                 & \multicolumn{1}{c|}{-}                                                                     & {78.35}                   & {83.14}                   & \multicolumn{1}{c|}{71.32}                   & {68.09}                   & {73.65}                   & \multicolumn{1}{c|}{58.03}                   & \multicolumn{1}{c|}{68.67}                   & -                                                                    \\
\multicolumn{1}{c|}{CM-A-E+Ref-NMS~\cite{chen2021ref}}             & \multicolumn{1}{c|}{R-101 \& LSTM}                                                                 & \multicolumn{1}{c|}{-}                                                                     & {\underline{80.70}}                   & {\underline{84.00}}                   & \multicolumn{1}{c|}{{\underline{76.04}}}                   & {\underline{68.25}}                   & {\underline{73.68}}                   & \multicolumn{1}{c|}{{\underline{59.42}}}                   & \multicolumn{1}{c|}{{\underline{70.62}}}                   & -                                                                    \\
\multicolumn{1}{c|}{NMTree~\cite{DBLP:conf/iccv/LiuZZW19}}                   & \multicolumn{1}{c|}{R-101 \& T-LSTM}                                                               & \multicolumn{1}{c|}{-}                                                                     & 76.41                         & 81.21                         & \multicolumn{1}{c|}{70.09}                         & 66.46                         & 72.02                         & \multicolumn{1}{c|}{57.52}                         & \multicolumn{1}{c|}{64.62}                         & -                                                                    \\ \hline \hline
\multicolumn{11}{c}{\textit{One-stage methods}}                                                                                                                                                                                                                                                                                                                                                                                                                                                                                                                                                                                     \\ \hline
\multicolumn{11}{l}{\textbf{\emph{CNN-based}}}                                                                                                                                                                                                                                                                                                                                                                                                                                                                                                                                                                                             \\ \hline
\multicolumn{1}{c|}{SSG~\cite{DBLP:journals/corr/abs-1812-03426}}                      & \multicolumn{1}{c|}{D-53 \& LSTM}                                                                 & \multicolumn{1}{c|}{416$\times$416}                                                                   & -                             & 76.51                         & \multicolumn{1}{c|}{67.50}                         & -                             & 62.14                         & \multicolumn{1}{c|}{49.27}                         & \multicolumn{1}{c|}{47.47}                         & 25                                                                   \\
\multicolumn{1}{c|}{FAQA~\cite{yang2019fast}}                     & \multicolumn{1}{c|}{D-53 \& BERT}                                                                 & \multicolumn{1}{c|}{256$\times$256}                                                                   & 72.54                         & 74.35                         & \multicolumn{1}{c|}{68.50}                         & 56.81                         & 60.23                         & \multicolumn{1}{c|}{49.60}                         & \multicolumn{1}{c|}{56.12}                         & 23                                                                   \\
\multicolumn{1}{c|}{RCCF~\cite{DBLP:conf/cvpr/LiaoLLWCQL20}}                     & \multicolumn{1}{c|}{DLA-34 \& LSTM}                                                                     & \multicolumn{1}{c|}{512$\times$512}                                                                   & -                             & 81.06                         & \multicolumn{1}{c|}{71.85}                         & -                             & 70.35                         & \multicolumn{1}{c|}{56.32}                         & \multicolumn{1}{c|}{-}                             & 25                                                                   \\
\multicolumn{1}{c|}{ReSC-Large~\cite{yang2020improving}}               & \multicolumn{1}{c|}{D-53 \& BERT}                                                                 & \multicolumn{1}{c|}{256$\times$256}                                                                   & 77.63                         & 80.45                         & \multicolumn{1}{c|}{72.30}                         & 63.59                         & 68.36                         & \multicolumn{1}{c|}{56.81}                         & \multicolumn{1}{c|}{63.12}                         & 36                                                                   \\
\multicolumn{1}{c|}{LBYLNet~\cite{huang2021look}}                  & \multicolumn{1}{c|}{D-53 \& BERT}                                                                 & \multicolumn{1}{c|}{256$\times$256}                                                                   & 79.67                         & 82.91                         & \multicolumn{1}{c|}{74.15}                         & 68.64                         & 73.38                & \multicolumn{1}{c|}{59.49}                         & \multicolumn{1}{c|}{62.70}                         & 30                                                                   \\
\multicolumn{1}{c|}{SOT-CNN(ours)}                    & \multicolumn{1}{c|}{D-53 \& BERT}                                                                 & \multicolumn{1}{c|}{256$\times$256}                                                                   & 81.35                & 82.82                         & \multicolumn{1}{c|}{77.35}                & 68.65                & 73.80                         & \multicolumn{1}{c|}{61.18}                & \multicolumn{1}{c|}{63.87}                & 31                                                                      \\ 
\multicolumn{1}{c|}{SOT-CNN(ours)}                    & \multicolumn{1}{c|}{D-53 \& BERT}                                                                 & \multicolumn{1}{c|}{416$\times$416}                                                                   & \textbf{82.11}                & \textbf{83.79}                         & \multicolumn{1}{c|}{\textbf{78.14}}                & \textbf{70.89}                & \textbf{75.81}                         & \multicolumn{1}{c|}{\textbf{62.75}}                & \multicolumn{1}{c|}{\textbf{65.29}}                & 39                                                                     \\
\hline
\multicolumn{11}{l}{\textbf{\emph{Transformer-based}}}                                                                                                                                                                                                                                                                                                                                                                                                                                                                                                                                                                                     \\ \hline
\multicolumn{1}{c|}{TransVG~\cite{deng2021transvg}}                  & \multicolumn{1}{c|}{R-50 \& BERT}                                                                  & \multicolumn{1}{c|}{640$\times$640}                                                                   & 80.32                         & 82.67                         & \multicolumn{1}{c|}{78.12}                         & 63.50                         & 68.15                         & \multicolumn{1}{c|}{55.63}                         & \multicolumn{1}{c|}{66.56}                         & 61.77                                                                \\
\multicolumn{1}{c|}{TransVG~\cite{deng2021transvg}}                  & \multicolumn{1}{c|}{R-101 \& BERT}                                                                 & \multicolumn{1}{c|}{640$\times$640}                                                                   & 81.02 & 82.72 & \multicolumn{1}{c|}{78.35} & 64.82 & 70.70 & \multicolumn{1}{c|}{56.94} & \multicolumn{1}{c|}{67.02} & -                                                                     \\
\multicolumn{1}{c|}{RefTrans$^\dagger$~\cite{li2021referring}}                  & \multicolumn{1}{c|}{R-101 \& BERT}                                                                 & \multicolumn{1}{c|}{640$\times$640}                                                                   & 82.23                        & 85.59                       & \multicolumn{1}{c|}{76.57}                         & 71.58                         & 75.96               & \multicolumn{1}{c|}{62.16}                        & \multicolumn{1}{c|}{69.41}                        & 41$^\star$ \\
\multicolumn{1}{c|}{SeqTR~\cite{zhu2022seqtr}}                  & \multicolumn{1}{c|}{D-53 \& GRU}                                                                 & \multicolumn{1}{c|}{640$\times$640}                                                                   & 83.72                         & 86.51                        & \multicolumn{1}{c|}{81.24}                         & 71.45                         & 76.26               & \multicolumn{1}{c|}{64.88}                        & \multicolumn{1}{c|}{71.50}                        & 50 
\\
\multicolumn{1}{c|}{SOT-Trans(ours)}                  & \multicolumn{1}{c|}{D-53 \& GRU}                                                                 & \multicolumn{1}{c|}{640$\times$640}                                                                   & {\textit{\textbf{84.86}}}                         & {\textit{\textbf{87.03}}}                         & \multicolumn{1}{c|}{\textit{\textbf{82.54}}}                         & {\textit{\textbf{71.69}}}                         & {\textit{\textbf{77.32}}}               & \multicolumn{1}{c|}{\textit{\textbf{65.92}}}                        & \multicolumn{1}{c|}{\textit{\textbf{72.67}}}                        & 52 
\\
\bottomrule
\end{tabular}}
% \vspace{-8pt}
\end{table*}

%------------------------------------------------------------------------
\section{Experiments}

%-------------------------------------------------------------------------
\label{subsec:experimental_setup}

\noindent \textbf{Datasets.}
We conduct experiments on four benchmark datasets, namely ReferIt~\cite{DBLP:conf/emnlp/KazemzadehOMB14}, RefCOCO~\cite{DBLP:conf/eccv/YuPYBB16}, RefCOCO+~\cite{DBLP:conf/eccv/YuPYBB16} and RefCOCOg~\cite{DBLP:conf/cvpr/MaoHTCY016}. Referred entities in ReferIt are selected from SAIAPR-12 dataset~\cite{DBLP:journals/cviu/EscalanteHGLMMSPG10}, where the entities can be objects or stuff (\textit{e.g.,} sky). There are 54,127/ 5,842/ 60,103 referring expressions in “train"/ “validation"/ “test" set respectively. For RefCOCO~\cite{DBLP:conf/eccv/YuPYBB16}, RefCOCO+~\cite{DBLP:conf/eccv/YuPYBB16} and RefCOCOg~\cite{DBLP:conf/cvpr/MaoHTCY016}, the objects are selected from MSCOCO~\cite{DBLP:conf/eccv/LinMBHPRDZ14}, hence in total there are  which 80 object categories. Both RefCOCO and RefCOCO+ are splited into “train", “validation", “testA" and “testB" set following~\cite{DBLP:conf/eccv/YuPYBB16}. Referred objects in “testA" are people, while those in “testB" are objects of other categories. “Train"/ “validation"/ “testA"/ “testB" has 120,624/ 10,834/ 5,657/ 5,095 referring expressions for RefCOCO, and 120,191/ 10,758/ 5,726/ 4,889 referring expressions for RefCOCO+. No position pointing words (\textit{e.g.,} “left", ”right") appeared in RefCOCO+, which increases the difficulty of locating the target object.  

\noindent \textbf{Implementation details.}
DarkNet-53 pre-trained on MSCOCO is used as the image encoder, and LSTM or BERT~\cite{DBLP:journals/corr/abs-1910-03771} is used as the text encoder. 
Since previous one-stage VG methods~\cite{DBLP:journals/corr/abs-1812-03426,yang2019fast,DBLP:conf/cvpr/LiaoLLWCQL20,deng2021transvg} adopt different input resolutions, we resize input image size to 256$\times$256, 416$\times$416 and 640$\times$640 to evaluate the performances under different resolutions. Ablations are conducted with input resolution 256$\times$256 unless specified. 
The Adam~\cite{kingma2014adam} optimizer with an initial learning rate of $1e^{-4}$, weight decay of $1e^{-4}$, and batch size of 64 is used to train our network. The learning rate is decreased with a cosine annealing strategy~\cite{DBLP:conf/iclr/LoshchilovH17}. Following previous work~\cite{huang2021look}, we train our network 100 epochs on ReferIt, RefCOCO and RefCOCO+, and 30 epochs on RefCOCOg.
During the evaluation, we deactivate our proposed exploration by random connection strategy to eliminate randomness. Following prior works~\cite{yang2019fast,yang2020improving,huang2021look}, Pr@0.5(\%) is adopted as the evaluation metric. Inference latency is tested on the Geforce-GTX-1080Ti GPU and CUDA 10.2 with Intel(R) Xeon(R) CPU E5-2640 v4 @ 2.40GHz.

\begin{table}[t]
\centering
\caption{Comparison with state-of-the-art methods on ReferIt.}% ``mix" indicates that~\cite{DBLP:conf/iccv/SadhuCN19} first resize the image to 300$\times$300 and later retrain with image sizes 600$\times$600.}
\label{SOTA:ReferIt}
\scalebox{0.9}
{\begin{tabular}{ccccc}
\toprule
\multicolumn{1}{c|}{Methods}        & \multicolumn{1}{c|}{\begin{tabular}[c]{@{}c@{}}Visual \& Text \\ Encoder\end{tabular}} & \multicolumn{1}{c|}{\begin{tabular}[c]{@{}c@{}}Input \\ size\end{tabular}} & \multicolumn{1}{c|}{\begin{tabular}[c]{@{}c@{}}ReferIt\\ test\end{tabular}} & \begin{tabular}[c]{@{}l@{}}Time\\ (ms)\end{tabular} \\ \hline
\multicolumn{5}{c}{\textit{Two-stage Methods}}                                                                                                                                                                                                                              \\ \hline
\multicolumn{1}{c|}{CMN~\cite{DBLP:conf/cvpr/HuRADS17}}            & \multicolumn{1}{c|}{VGG16 \& LSTM}                                                     & \multicolumn{1}{c|}{-}                                & \multicolumn{1}{c|}{28.33}        & -                                          \\
\multicolumn{1}{c|}{VC~\cite{DBLP:conf/cvpr/ZhangNC18}}             & \multicolumn{1}{c|}{VGG16 \& LSTM}                                                     & \multicolumn{1}{c|}{-}                               & \multicolumn{1}{c|}{31.13}        & -                                                \\
\multicolumn{1}{c|}{MAttNet~\cite{DBLP:conf/cvpr/Yu0SYLBB18}}        & \multicolumn{1}{c|}{R-101 \& LSTM}                                                & \multicolumn{1}{c|}{-}                                   & \multicolumn{1}{c|}{29.04}        & 184                                                  \\
\multicolumn{1}{c|}{Similarity Net~\cite{DBLP:journals/pami/WangLHL19}} & \multicolumn{1}{c|}{R-101 \& -}                                                   & \multicolumn{1}{c|}{-}                                & \multicolumn{1}{c|}{34.54}        & 196                                                  \\
\multicolumn{1}{c|}{CITE~\cite{DBLP:conf/eccv/PlummerKKZPL18}}           & \multicolumn{1}{c|}{R-101 \& -}                                                   & \multicolumn{1}{c|}{-}                               & \multicolumn{1}{c|}{35.07}        & 320                                            \\
\multicolumn{1}{c|}{DDPN~\cite{DBLP:conf/ijcai/YuYXZ0T18}}           & \multicolumn{1}{c|}{R-101 \& LSTM}                                                & \multicolumn{1}{c|}{-}                                   &\multicolumn{1}{c|}{\underline{63.00}}     & -                                              \\ \hline \hline
\multicolumn{5}{c}{\textit{One-stage Methods}}                                                                                                                                                                                                                              \\ \hline
\multicolumn{4}{l}{\textbf{\emph{CNN-based}}}                                                                                                                                                                                                                                  \\ \hline
\multicolumn{1}{c|}{SSG~\cite{DBLP:journals/corr/abs-1812-03426}}            & \multicolumn{1}{c|}{D-53 \& LSTM}                                                & \multicolumn{1}{c|}{416$\times$416}                    & \multicolumn{1}{c|}{54.24}         & 25                                                  \\
\multicolumn{1}{c|}{ZSGNet~\cite{DBLP:conf/iccv/SadhuCN19}}         & \multicolumn{1}{c|}{R-50 \& LSTM}                                                 & \multicolumn{1}{c|}{mix\tablefootnote{``mix'' means firstly train the model on the image that resized to 300$\times$300 and then retrain the model on images with the size of 600$\times$600.}}                                  & \multicolumn{1}{c|}{58.63}         & 25                                                  \\
\multicolumn{1}{c|}{FAQA~\cite{yang2019fast}}           & \multicolumn{1}{c|}{D-53 \& BERT}                                                & \multicolumn{1}{c|}{256$\times$256}                                         & \multicolumn{1}{c|}{60.67}         & 23                                                  \\
\multicolumn{1}{c|}{RCCF~\cite{DBLP:conf/cvpr/LiaoLLWCQL20}}           & \multicolumn{1}{c|}{DLA-34 \& LSTM}                                                    & \multicolumn{1}{c|}{512$\times$512}                    & \multicolumn{1}{c|}{63.79}         & 25                                          \\
\multicolumn{1}{c|}{ReSC-Large~\cite{yang2020improving}}     & \multicolumn{1}{c|}{D-53 \& BERT}                                                & \multicolumn{1}{c|}{256$\times$256}                                    & \multicolumn{1}{c|}{64.60}         & 36                                                  \\
\multicolumn{1}{c|}{LBYLNet~\cite{huang2021look}}        & \multicolumn{1}{c|}{D-53 \& BERT}                                                & \multicolumn{1}{c|}{256$\times$256}                                        & \multicolumn{1}{c|}{67.47}         & 30                                                  \\
% \multicolumn{1}{c|}{Ours1}          & \multicolumn{1}{c|}{D-53 \& LSTM}                                                & \multicolumn{1}{c|}{256$^{2}$}                                                   & 66.11                                                  \\
\multicolumn{1}{c|}{SOT-CNN(ours)}          & \multicolumn{1}{c|}{D-53 \& BERT}                                                & \multicolumn{1}{c|}{256$\times$256}                                                   & \multicolumn{1}{c|}{68.57}           & 31                                         \\ 
\multicolumn{1}{c|}{SOT-CNN(ours)}          & \multicolumn{1}{c|}{D-53 \& BERT}                                                & \multicolumn{1}{c|}{416$\times$416}                                                   & \multicolumn{1}{c|}{\textbf{69.50}}  & 39                                         \\
\hline
\multicolumn{5}{l}{\textbf{\emph{Transformer-based}}}                                                                                                                                                                                                                             \\ \hline
\multicolumn{1}{c|}{TransVG~\cite{deng2021transvg}}        & \multicolumn{1}{c|}{R-50 \& BERT}                                                 & \multicolumn{1}{c|}{640$\times$640}                                      & \multicolumn{1}{c|}{69.76}       & 62                                        \\
\multicolumn{1}{c|}{TransVG~\cite{deng2021transvg}}        & \multicolumn{1}{c|}{R-101 \& BERT}                                                & \multicolumn{1}{c|}{640$\times$640}                                      & \multicolumn{1}{c|}{70.73}   & -                        \\ 
\multicolumn{1}{c|}{RefTrans$^\dagger$~\cite{li2021referring}}        & \multicolumn{1}{c|}{R-101 \& BERT}                                                & \multicolumn{1}{c|}{640$\times$640}                                      & \multicolumn{1}{c|}{\textit{\textbf{71.42}}}   & -                        \\ 
\multicolumn{1}{c|}{Seq-TR~\cite{zhu2022seqtr}}        & \multicolumn{1}{c|}{D-53 \& GRU}                                                & \multicolumn{1}{c|}{640$\times$640}                                      & \multicolumn{1}{c|}{69.66}   & 50                        \\ 
\multicolumn{1}{c|}{SOT-Trans(ours)}        & \multicolumn{1}{c|}{D-53 \& GRU}                                                & \multicolumn{1}{c|}{640$\times$640}                                      & \multicolumn{1}{c|}{70.21}   & 52                        \\ 
\bottomrule
\end{tabular}}
% \vspace{-12pt}
\end{table}
% \vspace{-12pt}
\subsection{Comparison with State-of-the-art Methods}
\label{subsec:comparison_with_state_of_the_art_methods}
% \vspace{-4pt}
We compare our proposed method with state-of-the-art approaches on RefCOCO, RefCOCO+, RefCOCOg and ReferIt in Table~\ref{SOTA:COCO_series} and Table~\ref{SOTA:ReferIt}. For a more convenient comparison, we group existing methods into two-stage methods and one-stage methods, respectively. Besides, one-stage methods can be further split into CNN-based and Transformer-based methods. For fair comparisons, different encoders and different image resolutions are considered. 
%For model performance can be influenced by different encoders and input image sizes, we report the vision encoder, text encoder as well as image resolution of every method for fair comparisons. 

Table~\ref{SOTA:COCO_series} lists the results on COCO series datasets (\textit{i.e.,} RefCOCO, RefCOCO+, RefCOCOg). As is shown, our SOT-CNN  outperforms existing CNN-based one-stage methods in most cases. When using BERT as the text encoder, our method obtains 1.67\%, 3.2\% and 1.69\%  improvements on the RefCOCO val, testB set and RefCOCO+ testB set over the strong CNN-based competitor LBYLNet~\cite{huang2021look} when other settings keep the same.
Besides, our SOT-Trans also achieves non-trivial improvements over its baseline method SeqTR~\cite{zhu2022seqtr}, which is a strong transformer-based grounder as shown in Table~\ref{SOTA:COCO_series}. This demonstrates that the increased suspected object discrimination capability provided by our SOT also benefits when multi-modal information has been adequately integrated.

\begin{table}[!t]
    \centering
    \caption{Ablations for context modeling with different granularities in the one-stage visual grounding on the ReferIt dataset.}
    \label{abl:graph_type}
    \scalebox{1}
    {\begin{tabular}{l|c|c}
    \toprule
    \multicolumn{1}{c|}{Model} & \begin{tabular}[c]{@{}c@{}}Context Modeling \\ Granularities \end{tabular} & \begin{tabular}[c]{@{}c@{}}Pr@0.5\\ (\%)\end{tabular} \\ \hline
    \textit{Baseline}          & -                                                                         & 62.47                                                      \\ \hline
    +\textit{Nonlocal}         & \textit{dense grids}                                                 & 64.03                                                      \\
    +\textit{Deform Conv}      & \textit{dense grids}                                                 & 64.42                                                       \\
    +\textit{SOT}              & \textit{suspected objects}                                           & \textbf{66.11}                                                    \\
    \bottomrule
    \end{tabular}}
\end{table}

Table~\ref{SOTA:ReferIt} summarizes the performance comparisons on the ReferIt dataset. Generally, most of two-stage methods perform poorly on this dataset. This might be due to the poor candidate object detection results, since their detectors in the first stage are pre-trained on MSCOCO. As a result, no qualified candidate objects are available for matching with text queries. By getting rid of candidate object detection, one-stage methods, perform much better than most of two-stage methods. However, it is worth noting that the visual encoder (i.e., DarkNet-53) is also pretrained on MSCOCO, thus the model's initial selection of suspected objects is not reliable enough. As shown in Table~\ref{SOTA:ReferIt}, the consistent improvements over most CNN and transformer-based competitors prove the generalization capacity of our SOT.

\subsection{Ablation Studies}
\label{subsec:ablation_study}
% \vspace{-4pt}
We first conduct ablation studies on ReferIt dataset to verify the effectiveness of different modules in our framework, including sparse Suspected Object Transformation (SOT) for context modeling, Keyword-Aware Discrimination (KAD) and Exploration by Random Connection (ERC) for modulating model's initial choice on suspected objects. Here LSTM is used as the text encoder, and all models are trained with the same strategy described before for fair comparisons.

\noindent \textbf{Dense grids vs. sparse suspected objects.} Table~\ref{abl:graph_type} compares the proposed SOT against the basic one-stage grounder equipped with a non-local block~\cite{DBLP:conf/cvpr/0004GGH18} and a deformable convolution layer~\cite{DBLP:conf/iccv/DaiQXLZHW17}, two representative approaches to conduct exhaustive message passing among all grids. In the table, the \textit{Baseline} is implemented by combining ``Multi-modal Encoder" with ``Prediction" presented in Fig.\ref{framework}, hence no visual inter-object relationship modeling is applied for the baseline. For fair comparisons, we include a standard nonlocal block and a deformable convolution layer on dense grids for visual inter-object relationship modeling. From the results, we have the following observations. First, our SOT with selected sparse suspected objects performs much better than the baseline equipped with a dense grids modeling layer (\textit{i.e., }with a nonlocal block and a deformable convolution layer), showing the advantages of reasoning on a few suspected objects for visual grounding. This is mainly due to that SOT helps reduce the noise introduced by irrelevant regions. Second, compared to the baseline, adding either a dense grids modeling layer or our proposed SOT could achieve much better performances, which demonstrates the importance of inter-object relationship modeling.

\noindent \textbf{Effectiveness of KAD and ERC.} We further verify the effectiveness of our KAD and ERC in Table~\ref{abl:KC_RP}. From the results, KAD and ERC improve the base model 0.51\% and 1.05\%, respectively. By applying both KAD and ERC on the base model, the performance can be further improved to 66.11\%. The results demonstrate the effectiveness of our proposed KAD and ERC in distinguishing the referred object from multiple suspected objects. 

\begin{table}[!t]
    \centering
    \begin{minipage}{0.45\linewidth}
        \caption{Performances of alternative strategies of ERC.}
        \label{strategy_RP}
        \begin{tabular}{c|c}
            \toprule
            Strategy               & Pr@0.5(\%) \\ \hline
            \textit{original}               & 65.53      \\
            \textit{reverse}                & 65.41      \\
            \textit{average}                & 65.73      \\
            \textit{random}                 & 65.57      \\
            \textit{ERC(Ours)} & \textbf{66.11}      \\ \bottomrule
        \end{tabular}
    \end{minipage} \hfill
    \begin{minipage}{0.45\linewidth}
        \caption{Performances of alternative strategies of KAD.}
        \label{strategy_KC}
        \begin{tabular}{c|c}
            \toprule
            Strategy              & Pr@0.5(\%)     \\ \hline
            --                    & 64.99          \\
            \textit{sentence}     & 64.61          \\
            \textit{word average} & 64.45          \\
            \textit{KAD(Ours)}     & \textbf{66.11} \\ \bottomrule
        \end{tabular}
    \end{minipage}
\end{table}

\begin{table}[!t]
    \centering
    \caption{Ablations for KAD or ERC on ReferIt dataset. ``w/o ERC" means calculating edge weights with Eq.(\ref{edge}).}
    \label{abl:KC_RP}
    \scalebox{1}
    {\begin{tabular}{cc|cc|c}
    \toprule
    \multicolumn{2}{c|}{Objects Modulation} & \multicolumn{2}{c|}{Relations Modulation} & \multirow{2}{*}{\begin{tabular}[c]{@{}c@{}}Pr@0.5\\ (\%)\end{tabular}} \\
    w/o KAD              & w/ KAD              & w/o ERC              & w/ ERC              &                                                                        \\ \hline
    \checkmark          &                   & \checkmark          &                   & 64.48                                                                       \\
    \checkmark          &                   &                     & \checkmark        & 64.99                                                                       \\
                        & \checkmark        & \checkmark          &                   & 65.53                                                                       \\
                        & \checkmark        &                     & \checkmark        & \textbf{66.11}                                                                       \\ 
    \bottomrule
    \end{tabular}}
% \vspace{-12pt}
\end{table}

\subsection{Alternative Strategies for ERC and KAD.}
\label{subsec:alternative_strategies}
% \vspace{-4pt}
To further verify the effectiveness of our proposed ERC and KAD, we demonstrate some other possible strategies for replacing them in Table~\ref{strategy_RP} and Table~\ref{strategy_KC}.

\noindent \textbf{ERC.} As for the ERC, there are some alternatives to calculate $\alpha'_{j}$ in Eq.(\ref{edge_exploration}) as listed below:
\begin{itemize}
    \item \textit{original}: $\alpha'_{j}=\alpha_{j}$. Keep $\alpha$ values unchanged. 
    \item \textit{reverse}: $\alpha'_{j}=\alpha_{K-j+1}$. Reverse descending sequence $(\alpha_{1},\dots,\alpha_{K})$ as an ascending sequence $(\alpha_{K},\dots,\alpha_{1})$.
    \item \textit{average}: $\alpha'_{j}=\frac{1}{K}\sum_{k=1}^{K}\alpha_{k}$. Average $\textit{K}$ activation scores, and regard the average value as $\alpha'_{j}$.
    \item \textit{random}: $\alpha'_{j}\sim \mathcal{N}(0,1)$. Randomly sample $\alpha$ values from the standard normal distribution.
\end{itemize}

 As shown in the Table~\ref{strategy_RP}, among five strategies, \textit{reverse} performs the worst as shown in Table~\ref{strategy_RP}, which  may result from that some model's initial correct judgements are entirely not trusted. As the most direct ways to eliminate the influence of model's confidence, \textit{average} and \textit{random} erases model's preference to $\textit{K}$ suspected objects with averaging and randomly sampling operations in repectively, thereby providing another chance for model to rethink. Although only achieving slight improvements (\textit{i.e.,} 0.2\% with \textit{average} and 0.04\% with \textit{random}), such results preliminarily validate the effectiveness of the idea that decoupling edge weights from model confidence  enables the model to rethink. 
Compared with other strategies, our proposed ERC not only complies with model's initial choice (\textit{i.e.,} keeping $\alpha'_{j}=\alpha_{j}$ with probability $p$), but also has chances to break up the dependencies between edge weights and model confidence (\textit{i.e.,} assigning $\alpha'_{j}$ with a value uniformly sampled from $\{\alpha_k\}_{k \in \{1,\dots,K\}\setminus\{j\}}$
with probability 1-$p$). With the above merits, our ERC achieves 0.58\% performance improvement compared with the baseline (\textit{i.e., original}) without introducing extra computation. 

\noindent \textbf{KAD.} 
As for the KAD, we design two counterparts (\textit{i.e., sentence} and \textit{word average}) as shown in Table~\ref{strategy_KC}, where the suspected object features are modulated by sentence-level feature from LSTM and averaged words feature, respectively. “-" in the first row means no language information is used to modulate suspected object features. We observe that simply using the sentence feature or the averaged words feature even degenerates the model, as they introduce too much noise from the irrelevant words. By excavating keywords, our KAD can bring 1.12\% absolute improvements on the ReferIt test set.

\begin{table}[t]
    \centering
    \caption{Comparison of mAT improvements brought by KAD and ERC.}
    \label{tab:mAT}
    \begin{tabular}{cc|c} \toprule
    KAD & ERC & mAT \\
    \hline
         &    & 50.82 \\
    \checkmark & & 56.02 (\textcolor{red}{$\uparrow$ 5.20}) \\
     & \checkmark &54.43 (\textcolor{red}{$\uparrow$ 3.61}) \\
     \checkmark & \checkmark & 59.08 (\textcolor{red}{$\uparrow$ 8.26}) \\
     \bottomrule
    \end{tabular}
% \vspace{-6pt}
\end{table}

\subsection{Suspected Objects Discrimination Ability Evaluation}
To quantitatively examine the improvement of suspected objects' discrimination ability brought by our SOT, we define a new evaluation metric dubbed ``Attentive Ratio" (AT) as:
\begin{equation}
\label{eq:at}
    \mathrm{AT} = {\sum_{v_m \cap B_{GT} = v_m} \alpha_m} / {\sum_{n \in \{1,2,\dots,K\}} \alpha_n}
\end{equation}
Recall that $\alpha_m$ is the activation strength of suspected objects $v_m$, and there are $K$ selected ones in total. $B_{GT}$ denotes the area of the ground-truth bounding box. 
In Eq.(\ref{eq:at}), the numerator calculates the total activation scores of suspected objects falling inside the ground-truth bounding box, and the denominator calculates total activation scores of $K$ suspected objects. Hence, the higher AT score represents higher attention paid to the target object, proving stronger suspected objects' discrimination ability. We ablate SOT's main components, KAD and ERC, by calculating the mAT value (mean of AT across all samples) as shown in the Table~\ref{tab:mAT}. Adding KAD or ERC can boost 5.20\% or 3.61\% mAT, respectively. And combination of them can further increase mAT from 50.82\% to 59.08\%. Therefore, our SOT with KAD and ERC designs can promote the model's visual grounding performance by enhancing the suspected objects' discrimination ability.

\section{Conclusions}
\label{sec:conclusions}
% \vspace{-4pt}
In this paper, we focused on the relation modeling of suspected objects in the one-stage visual grounding and proposed a Suspected Object Transformation (SOT) module. By adaptively selecting suspected objects and modulating them with Keyword-Aware Discrimination (KAD) and Exploration by Random Connection (ERC) mechanisms during training, our SOT could filter cluttered backgrounds and focus on discriminating the target objects from multiple suspected ones. 

\section{Acknowledgement}
This work was supported by the NSFC project (No. 62072116) and in part by Shanghai Science and Technology Program (No. 21JC1400600).

\vfill\eject
\bibliographystyle{ACM-Reference-Format}
\balance
\bibliography{cite}

\end{document}

\title{Suspected Objects Matter: Rethinking Model's Prediction for One-stage Visual Grounding -- Supplementary Materials}


\maketitle

%------------------------------------------------------------------------
\section{Visualization Analysis}
\label{sec:more_qualitative_results}

\begin{figure*}[!h]
    \centering
    \includegraphics[width=0.8\linewidth]{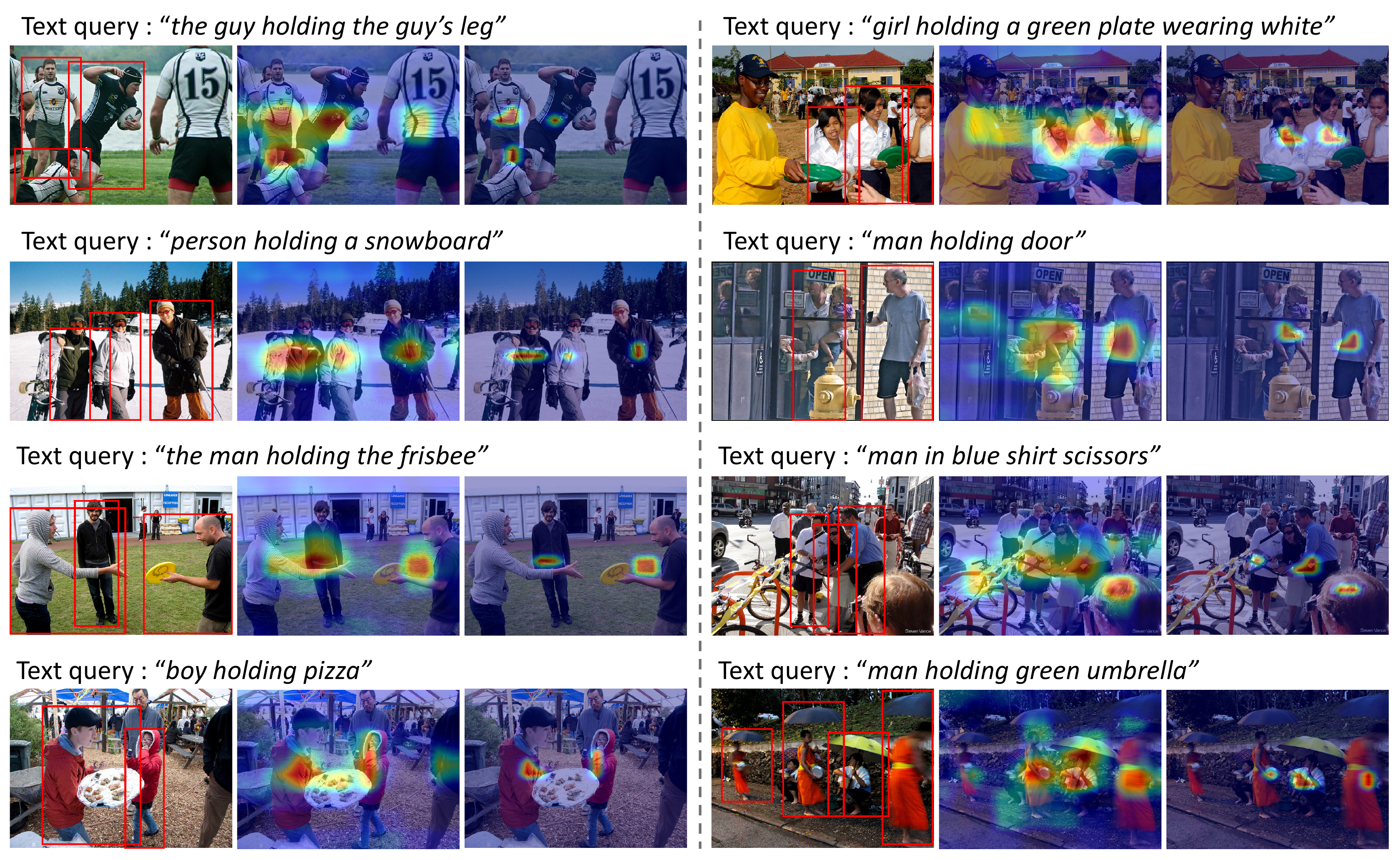}
    \caption{More visualization results to prove the consistency of suspected objects (labeled by red bounding boxes) and a naive one-stage grounder's activation distribution after training five epochs (the second picture within each column). We also illustrate selected suspected objects' area across each image at present (the last picture within each column).}
    \label{matching}
    % \vspace{-18pt}
\end{figure*}

\subsection{Consistency of model's initial confidence and suspected objects}
As we claimed in the main paper that based on the observation that the model's initial confidence (represented by the activation distribution) is consistent with the suspected objects, we can select the model's high activation areas as the regions including suspected objects. In this section, to give more supporting cues of such observation, we visualize more relevant cases as shown in Fig.\ref{matching}. Each case comprises three pictures, where the first one is the original image with suspected objects manually labeled by red bounding boxes, and the second and third ones are the activation map of a naive one-stage grounder and suspected objects selected according to this activation map, respectively. From these visualization results, we can observe that the model is able to pay high attention to instances highly consistent with the manually labeled suspected objects.

\begin{figure*}[!h]
    \centering
    \includegraphics[width=\linewidth]{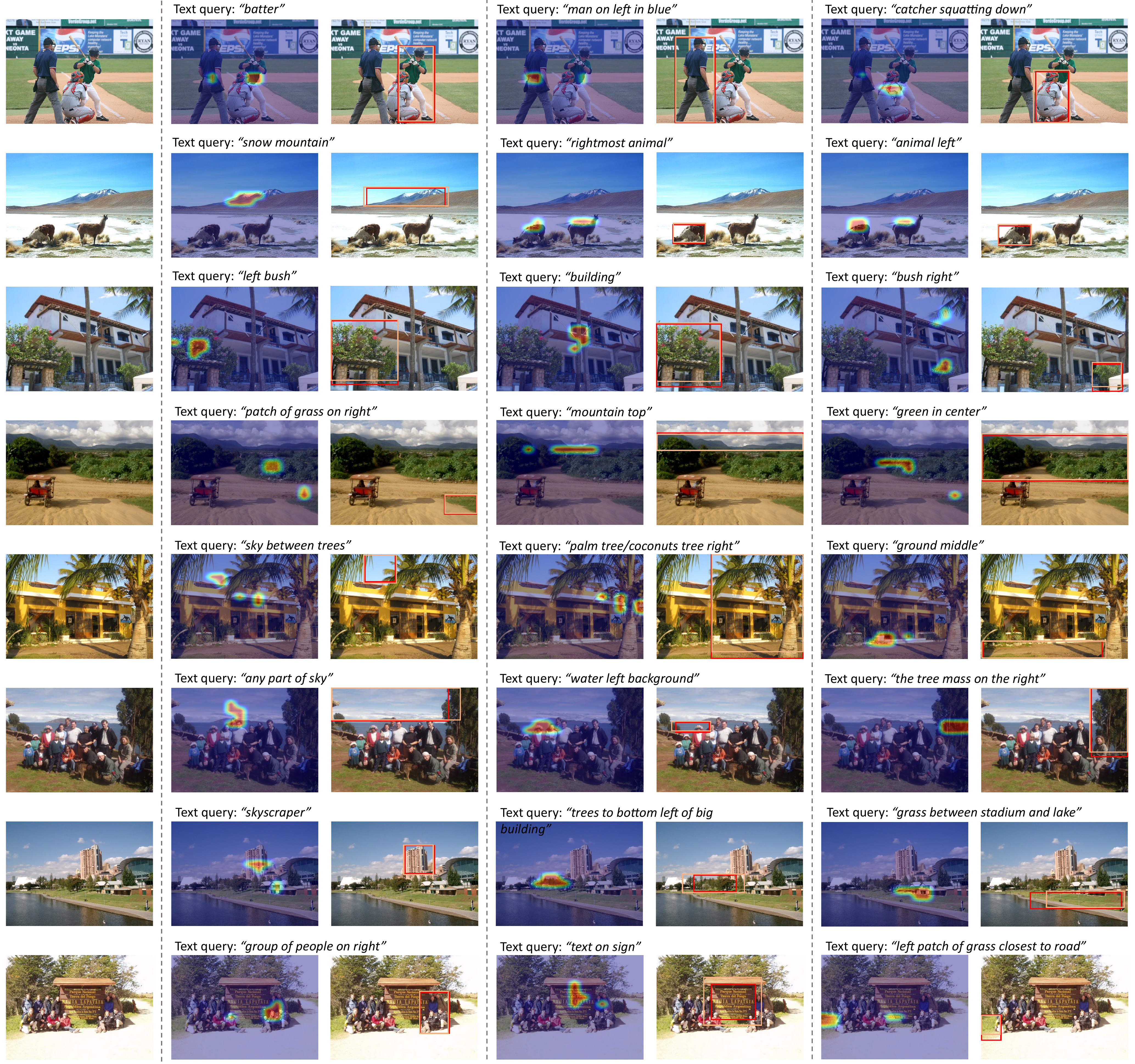}
    \caption{More qualitative results of our proposed SOT. Heat maps are generated by mapping the selected suspected objects' activation into the original image. Orange and red boxes are the prediction results and ground truths, respectively.}
    \label{more_qualitative_results}
    \vspace{-18pt}
\end{figure*}

\subsection{More qualitative results}
We provide more qualitative results in Fig.\ref{more_qualitative_results}, where each row includes three different language expressions for different referred objects on the same image. From the results, we observe that in most cases, selected suspected objects can converge to different objects conditioned on different language expressions. And with more selected grids attending to the referred objects, the whole model can generate prediction results highly consistent with the ground truths.

\bibliographystyle{ACM-Reference-Format}
%% \bibliography{sample-base}
\bibliography{cite}